\documentclass{article}
\usepackage{spconf,amsmath,epsfig}
\usepackage{amsthm,amsmath,amssymb}
\usepackage{multirow}
\usepackage{mathrsfs}
\usepackage{color}
\usepackage{url}
\let\OLDthebibliography\thebibliography
\renewcommand\thebibliography[1]{
  \OLDthebibliography{#1}
  \setlength{\parskip}{0pt}
  \setlength{\itemsep}{0pt plus 0.3ex}
}
\newcommand{\eat}[1]{}

\newcommand{\ie}{\textit{i}.\textit{e}., }
\newcommand{\eg}{\textit{e}.\textit{g}., }

\begin{document}\sloppy
\topmargin=0mm

% Example definitions.
% --------------------
% \def\x{{\mathbf x}}
% \def\L{{\cal L}}

% Title.
% ------
\title{Learning To Generate Scene Graph from Head to Tail}
%
% Single address.
% ---------------
% \name{Anonymous ICME submission}
% %Address and e-mail should NOT be added in the submission paper. They should be present only in the camera ready paper. 
% \address{}
\name{Chaofan Zheng$^{1,2}$, Xinyu Lyu$^{1}$, Yuyu Guo$^{1}$, Pengpeng Zeng$^{1,2}$, Jingkuan Song$^{1}$, Lianli Gao$^{1,*}{\thanks{* Corresponding author.}}$
\thanks{This work is supported by National Key Research and Development Program of China (No. 2018AAA0102200), the National Natural Science Foundation of China (Grant No. 62122018, No. 61772116, No. 61872064), Sichuan Science and Technology Program (Grant No.2019JDTD0005, No.2022119).}}
% \address{$^{\ast}$First author address and e-mail; $^{\dagger}$Second author address and e-mail ; (...) and \\ $^{\ddagger}$Last author address and e-mail.}
\address{$^{1}$University of Electronic Science and Technology of China, China\\
$^{2}$Sichuan Artificial Intelligence Research Institute, Yibin, China\\
\{chaofan.zheng99, xinyulyu68, yuyuguo1994, is.pengpengzeng, jingkuan.song\}@gmail.com \\
lianli.gao@uestc.edu.cn}

\maketitle

% Scene graph generation (SGG) represents objects and their interactions with a graph structure, which plays a vital role in image understanding tasks. 
% Although many works are devoted to generating an unbiased scene graph, underestimating the head predicates in the whole training process, they wreck features of head predicates that provide fundamental features for tail predicates. 
% Based on this observation, we propose an SGG framework, learning to generate scene graph from Head to Tail (SGG-HT), which contains Curriculum Re-weight Strategy and Semantic Context Module. 
% The Curriculum Re-weight Strategy learns head/easy predicates at first for robust features of head samples and then gradually focuses on tail/hard ones. 
% Besides, The Semantic Context Module is proposed to relieve semantic errors caused by assigning excessive weight to the tail predicates.
% Extensive experiments show that our method significantly alleviates unbiased problem and achieves state-of-the-art performance on the Visual Genome dataset.

\begin{abstract}

Scene Graph Generation (SGG) represents objects and their interactions with a graph structure. 
Recently, many works are devoted to solving the imbalanced problem in SGG. 
However, underestimating the head predicates in the whole training process, they wreck the features of head predicates that provide general features for tail ones.
Besides, assigning excessive attention to the tail predicates leads to semantic deviation. 
Based on this, we propose a novel SGG framework, learning to generate scene graphs from Head to Tail \textbf{(SGG-HT)}, containing \textbf{Curriculum Re-weight Mechanism (CRM)} and \textbf{Semantic Context Module (SCM)}. CRM learns head/easy samples firstly for robust features of head predicates and then gradually focuses on tail/hard ones. SCM is proposed to relieve semantic deviation by ensuring the semantic consistency between the generated scene graph and the ground truth in global and local representations.
Experiments show that SGG-HT significantly alleviates the biased problem and achieves state-of-the-art performances on Visual Genome.

% 目前许多工作致力于解决场景图生成中的长尾分布问题，但是他们存在了一些问题：First,  然而低估了头部谓词在整个训练过程中，他们损害了头部谓词的特征that：， 此外由于过度关注于尾部谓词，其又会在尾部谓词上过拟合。 

% Recently, many works are devoted to solving imbalanced problem in scene graph generation.However, underestimating the head predicates in the whole training process, they wreck the features of head predicates that provide general features for tail ones. Besides, assigning excessive attention to the tail predicates leads to tail overfitting. 

\end{abstract}

\section{Introduction}
\label{sec:intro}

% 问题背景（为什么研究的问题重要）
% 问题的挑战（研究的问题很有难度）
% 现有方法还没有解决的很好（自己的机会）
% 提出的方法（核心idea）
% 总结方法的Novelty

% 场景图的应用
Scene Graph Generation is a fundamental task of computer vision that involves detecting the objects and their relationships in an image to generate a graph structure. Such a structured representation is helpful for downstream tasks such as Visual Question Answering~\cite{vqa} and Image Captioning~\cite{imgcap}. 

% 场景图的一般做法以及存在的问题
% Existing models for scene graph generation~\cite{sgg:motifs,sgg:graphrcnn,sgg:bgnn,sgg:vctree} typically start with an object detection network, which generates a set of proposals and corresponding features. Then, a contextual message passing module is used to refine the visual features. Finally, these refined features are used to classify predicate labels. 
% These methods achieve excellent performances on the \textbf{Recall} metric. However, for the balanced metric, \ie \textbf{Mean Recall}~\cite{sgg:kern, sgg:vctree}, their performances are disappointed due to the long-tailed data distribution. 
% % that a few predicates have abundant samples while most predicates only have a few samples.
% Specially, the predictions of previous models are dominated by the head predicates because of the extremely biased data distribution, \eg falsely predicting ``on" instead of ``riding", and coarsely predicting ``on" instead of ``sitting on"

Although previous works~\cite{sgg:motifs,sgg:vctree, sgg:graphrcnn} make great efforts to improve the context aggregation capacity of the model, their performances are disappointed due to the long-tailed data distribution. Especially, the predictions of previous models are dominated by the head predicates, \eg falsely predicting ``on" instead of ``riding" and coarsely predicting ``on" instead of ``sitting on". 
To solve this problem, ~\cite{sgg:pcpl} proposes a novel re-weighting method that utilizes the correlation among predicate classes to seek out appropriate loss weights adaptively. ~\cite{sgg:cogtree} builds a hierarchical cognitive structure from the cognition perspective to make the tail relationships receive more attention in a coarse-to-fine mode.

\begin{figure}[t]
  \center
  \includegraphics[width=1\linewidth]{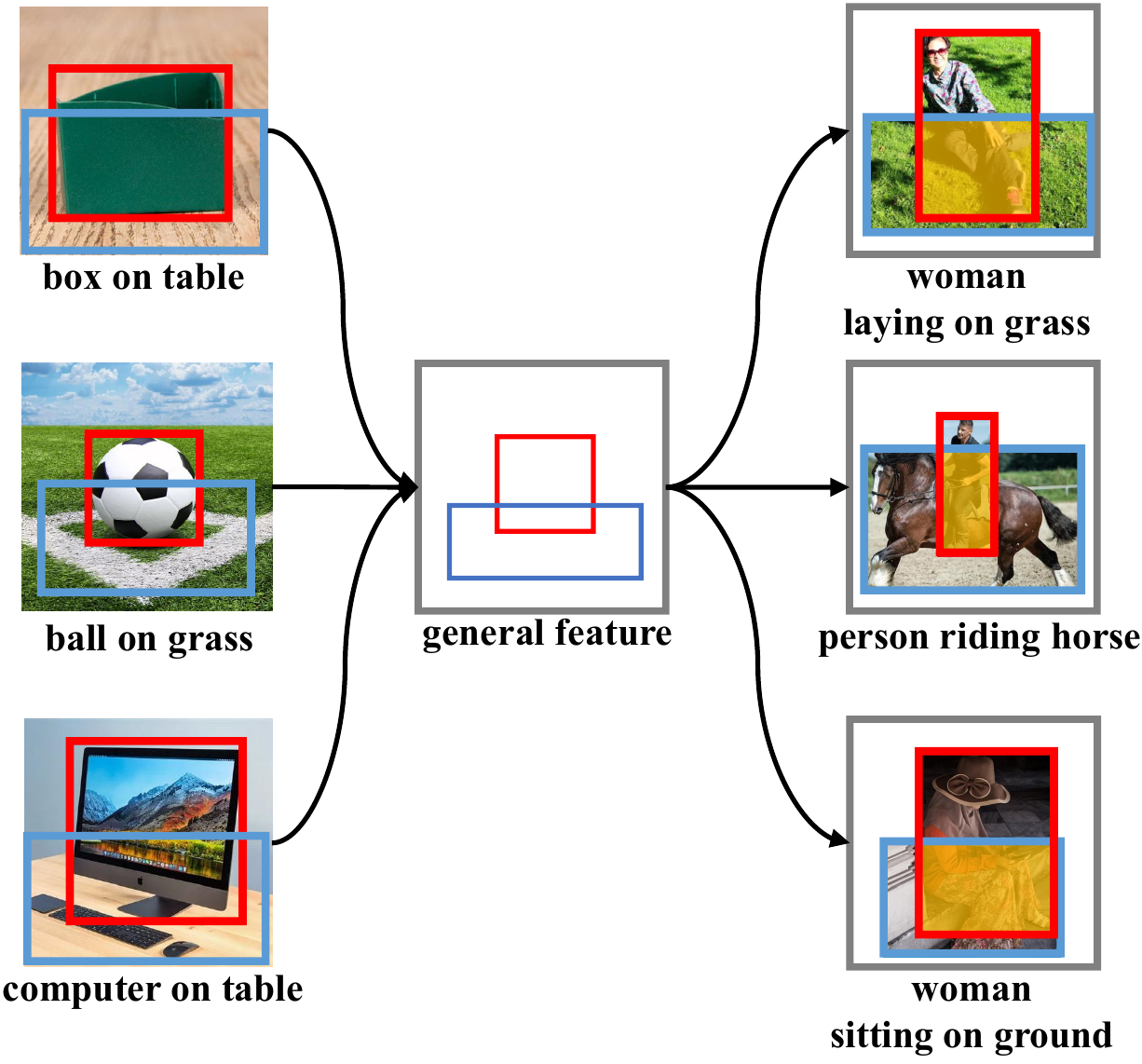}
  \vspace{-22pt}
\caption{Head predicates can provide general features for tail predicates. 
For instance, the predicate ``on" describes an object on top of another object.
When different contexts are added based on this general feature, different tail predicate features are generated.
% This general feature of ``on'' may benefit the learning of some tail predicates, \eg ``lying on", ``sitting on" and "riding" . 
}
\vspace{-14pt}

\label{fig:intro}
\end{figure}

However, underestimating the head predicates during training, these debiasing methods wreck the features of head predicates. Moreover, since the features of tail predicates may depend on those of head ones, \eg the feature of ``sitting on'' depends on that of ``on'', 
% head predicate "on" benefits the learning of its correlated tail predicates, e.g., "lying on", "sitting on" and "riding", sharing fundamental features that one object is  on top of the other, 
these wrecked features of head predicates also harm the learning of tail ones. Therefore, handling the biased problem in SGG requires the robustness of head predicates firstly.
As shown in Fig.~\ref{fig:intro}, the head predicate ``on" may benefit the learning of its correlated tail predicates, \eg ``lying on", ``sitting on" and ``riding", sharing general features that one object is on top of the other. 
% By weighting the head predicates with minor values in the whole training, these debiased methods focus on tail predicates but impair the learning of head predicates. 
% However, the learning quality of tail predicates is positively correlated with head ones. As shown in Figure \ref{fig:fig1}, head predicate ``on" can provide a fundamental feature, \ie an object on top of another one, and the tail predicates ``lying on", ``sitting on" and ``riding" depend on the fundamental feature of ``on''. Therefore, learning tail predicates requires robust features of head predicates as a basis. However, these methods that wreck the features of head predicates damage the learning of tail predicates.
Besides, they ignore the semantic deviation problem caused by assigning excessive weights to tail predicates that may incorrectly predict a head predicate as an unrelated tail predicate, \eg wrongly predicting the ``on" to ``using". Thus, their performances are sub-optimal.
% Because these methods damage the features of head predicates,  
% ignore the importance and generality of the features of head predicates that the head predicates can provide class-generic features for tail predicates, as shown in Figure \ref{fig:fig1}. 
% Besides, they also ignore the empirical prediction problem caused by giving the tail predicates a over-high weight, e.g. wrongly predicting the  ``on" to ``eating". Thus, their performance is sub-optimal. 

% 这些方法破坏了头部谓词的学习。
% 然而，尾部谓词依赖于头部谓词的学习。如图1所示. XXX
% 所以尾部谓词的学习需要一个鲁棒的头部谓词的特征。
% 因为他们这些方法损害了头部谓词的学习，以至于得到了一个被破坏的头部谓词的特征，因此尾部谓词也没有得到一个很好的学习。

% 语义大的误差，比如在两个语义无关上的混淆。
% 提出自己的方法
In this paper, we propose an SGG framework, learning to generate scene graphs from Head to Tail (SGG-HT), consisting of \textbf{Curriculum Re-weight Mechanism} and \textbf{Semantic Context Module}. 
The \textbf{Curriculum Re-weight Mechanism} adjusts relative weights between head and tail predicates through a curriculum decay factor. It regulates the model to learn the robust features of the head predicates at first and then gradually focus on the tail ones to better use the general features provided by the head predicates.
% This learning process from easy to hard effectively improves the performance.
\textbf{Semantic Context Module} takes the local semantic representations of relation triplets and the global semantic representation of the whole graph as inputs and generates the contextual semantic representations. 
The global contextual representation is used to measure the overall semantic gap between the generated scene graph and the ground truth. 
The local contextual representations can correct mispredictions in each triplet individually.
With the Semantic Context Module, the model can alleviate the problem of semantic deviation and generate a scene graph consistent with the actual semantics, \ie the semantics of the ground truth scene graph.

% 使用global representation直接的确保语义相似，其余的用于更正原始的预测，间接的确保语义相似
% to correct the original predicates predictions. Besides, we design a global representation of the scene graph of an image, 
% and use it to ensure the consistence between the generated and real scene graph. With the Semantic Transformer, we can alleviate the problem of empirical prediction and generate scene graph that is more consistent with the real semantics.

% 对于ST，其将一张图片中的三元组的语义表征和一个额外的全局表征作为输入，然后生成具有上下文的语义表征，其中相应的三元组的语义表征被用来，全局的语义表征用于与真实场景图的全局语义表征进行对比。

% it takes the word embedding vectors of all the relation triplet labels in an image as inputs, and generates a global semantic representation vector as the semantic representation of an image scene. It uses the semantic representation of the real scene as supervision to effectively ensure the semantic consistency of the generated scene graph. Specially, it can also explore the dependencies between relation triples to generate context-aware relation semantic vectors. These context-aware relation semantic vectors fuse the information of the remaining relation triples, and can be used to modify the predictions of the model. 

% 对于semantic transformer，以语义表示和一个全局的语义表示作为输入，然后生成具有上下文的语义表征。
% 其中contextual global semantic被用于度量与真实语义之间的gap，其余的用于修改可能存在的错误预测

% 本文的贡献
Our proposed method is model-agnostic so that it can be integrated with most existing models. We extensively validate our method with different models. The experiments results show that our approach significantly improves the performance and consistently achieves state-of-the-art performance. The main contributions of our works are three-folds:
\begin{itemize}
    \vspace{-5pt}
      \item We propose a Curriculum Re-weight Mechanism, which benefits  the learning of tail predicates by taking advantage of the general features from the head predicates.
%   make the tail predicates use the fundamental features of the head ones for better learning.
    \vspace{-5pt}
%   it uses the inherent properties of predicate features that head predicates could provide class-generic feature for the tail predicates and makes a better performance.
    \item We propose a Semantic Context Module to alleviate the semantic deviation problem, which can make the semantics of the generated scene graph closer to the actual semantics.
  \vspace{-5pt}
%   by utilizing the semantic representations of relation triplets to ensure consistency of the semantic between the generated scene graph and the real ones.
     \item Our method achieves state-of-the-art performances with different models on Visual Genome. For instance, our method improves Motif~\cite{sgg:motifs} from 16.08 to 39.43, VCTree~\cite{sgg:vctree} from 18.16 to 40.21 and Transformer~\cite{sgg:sggbenchmark,transformer} from 17.63 to 42.60 on Predcls mR@100.
  
\end{itemize}
\vspace{-13pt}
\section{RELATED WORK}
\vspace{-3pt}
\subsection{Scene Graph Generation}
Scene Graph Generation (SGG) is a task that takes an image as input and generates a structured graph. 
Early work~\cite{sgg:vrd} detects objects and relationships via   independent networks. Subsequently, ~\cite{sgg:imp} proposed a message passing model to aggregate context information for reasoning.  
Afterward, other architectures~\cite{sgg:graphrcnn,sgg:motifs,sgg:vctree, sgg:kern} were designed to improve the context aggregation capacity for better performance.
Recently, ~\cite{sgg:kern} and~\cite{sgg:vctree} both notice the imbalanced problem in SGG and propose balanced metric Mean Recall@K. 
~\cite{sgg:tde} solves this problem by employing causal inference in the inference stage to remove the bad bias.
~\cite{sgg:ba-sgg} designs a framework to adjust the learning of the model from semantics and sample spaces to generate scene graphs with rich information. 
Our method generates an unbiased scene graph by providing the robust features of head predicates for tail ones and ensuring semantic consistency between the generated scene graph and the ground truth.

% Different from these works, we explore the inherent properties between predicates and the semantic consistency between generated scene graph and real scene graph to generate an unbiased scene graph. Our work base on Scene Graph Generation Codebase~\cite{sgg:sggbenchmark}.
    
% Our work considers the adaptability of the head predicate features and devises a simple learning strategy to help the complex tail predicate use these fundamental features. Besides, we also explore the semantic consistency of the scene graph and the correlation between the relation triplets. 

\vspace{-7pt}
\subsection{Long-Tailed Classification}
The data in the real world has a natural long-tailed distribution: a few categories (head class) have abundant samples while the majority categories (tail class) only have a few samples. 
A classic method to deal with long-tailed distribution is re-sampling. It makes the data distribution balanced, including oversampling for the tail classes~\cite{oversamp1, oversamp2} and undersampling for the head classes~\cite{undersamp2}. 
Another effective method is to re-weight the loss function~\cite{loss:cbloss,loss:focalloss}, which assigns different weights for different classes to balance the loss. 
% Focal Loss~\cite{loss:focalloss} reduces the weight of well-classified simple samples and focuses on the hard samples. Class-Balanced Loss~\cite{loss:cbloss} proposes the definition of effective number of samples and inverse of the effective number of each category as its weight to address the problem. 
However, ignoring the correlation between predicates that head predicates can provide general features for tail ones, these methods are not entirely appropriate for the scene graph generation.

% Although these methods are effective in some areas, they are not entirely appropriate for scene graph generation. because they ignore some inherent correlations between predicates.

\begin{figure*}[t]
 \centering
 \includegraphics[width=1.0\linewidth]{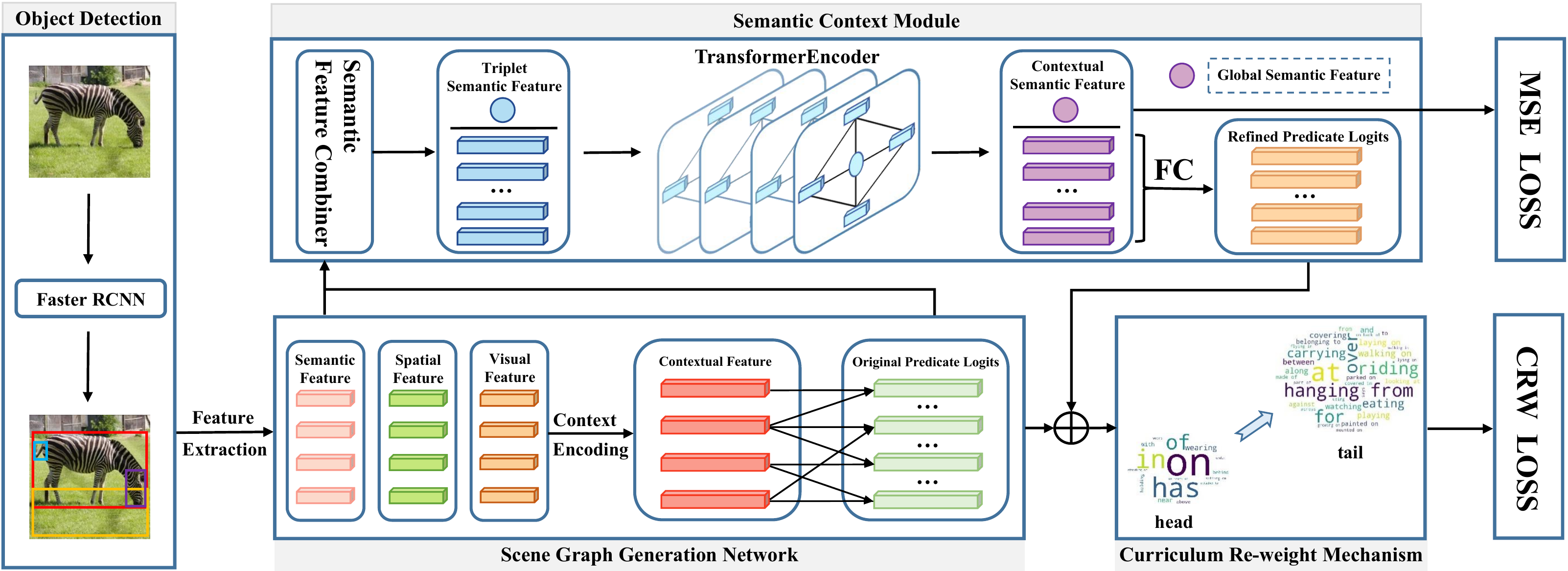} 
 \vspace{-17pt}
 \caption{Overview of the SGG-HT framework. 
The basic SGG framework adopts Faster RCNN to obtain spatial, visual, and semantic features. Next, a context encoding module is used to generate the contextual features for predicate classification. Our proposed SGG-HT framework includes two new components: (1) a Curriculum Re-weight Mechanism that gradually transfers the learning focus from head predicates to tail ones with Curriculum Re-Weight Loss; (2) a Semantic Context Module that relieves the semantic deviation between the generated scene graph and the ground truth. 
 }
 \vspace{-13pt}
 \label{fig:kjt}
\end{figure*}
\vspace{-4pt}
\section{METHOD}
\vspace{-4pt}
\label{sec.method}
We first give the problem definition of SGG in Sec.~\ref{sec:prob-def}. Then, a method is proposed to generate an unbiased scene graph, as illustrated in Fig.~\ref{fig:kjt}. In particular, our method consists of two components: a Curriculum Re-weight Mechanism in Sec.~\ref{sec:CRM}. and a Semantic Context Module in Sec.~\ref{sec:SCM}. 

\subsection{Problem definition}
\label{sec:prob-def}
Scene graph generation is a two-stage classification task. In the first stage, the Faster RCNN~\cite{faster-rcnn} framework is used to obtain object features for each image, including:
\begin{itemize}
    \vspace{-3pt}
    \item A set of spatial features $ B=\{b_1, b_2, ... , b_n\} $, where $b_i \in \mathbb{R}^4$  denotes the the spatial locations of detected regions.
    \vspace{-5pt}
    \item a set of region proposals' visual features $ V = \{v_1, v_2, ..., v_n\} $, where $v_i \in \mathbb{R}^{4096}$.  
    \vspace{-5pt}
    \item A set of object labels $ L = \{l_1, l_2, ... , l_n\}$, where $l_i \in \mathbb{R}^{O+1}$, $O$ is the number of object classes. Then we obtain the semantic features $ S^o = \{s_1^o, s_2^o, ... , s_n^o\}$ by mapping the object label $l_i$ to a 200 dimensional vector with a word embedding model.
    \vspace{-5pt}
\end{itemize}
% Suppose there are $O$ object classes and $R$ predicate classes. The object feature for the $i$-th proposal $x_i$ includes the visual feature $v_i \in \mathbb{R}^{4096}$, the spatial feature $b_i \in \mathbb{R}^4$ and the semantic feature $s_i^o \in \mathbb{R}^{200}$ which is the word embedding feature of the label probability $l_i \in \mathbb{R}^{O+1}$. 
In the second stage, the object features are refined by a context encoding module, such as BiLSTM~\cite{sgg:motifs}, GCN~\cite{sgg:graphrcnn}, or TreeLSTM~\cite{sgg:vctree}. Finally,
possible object pairs' contextual features are fed into a classifier to predict the predicate probability $p_i \in \mathbb{R}^{R+1}$.
% the contextual object features of possible object pairs are fed into a classifier 

Generally, the standard cross-entropy loss is used for predicate classification in the optimization process of the above methods. Given the predicate predicted logits $z= [z_1,z_2,...,z_{R+1}]$ ($R$ predicate classes and a background class) and ground truth label $y=[y_1,y_2,...,y_{R+1}]$, in which $y_i$ equals 1 or 0, the cross-entropy loss is formed as:
\vspace{-6pt}
\begin{equation}\label{Eq:3.1}
  \mathcal{L}_{CE}(z,y) =  -  \sum_{i=1}^{R+1} y_i \log{\frac{e^{z_i}}{\sum_{j=1}^{R+1} e^{z_j}}}.
    \vspace{-2pt}
\end{equation}
Due to the long-tailed data distribution, the predicates of tail classes have less frequency of occurrence. With the optimization of this loss function, the model's predictions will be dominated by head predicates. Thus, the classifier tends to predict biased classification scores resulting in low accuracy of tail predicates.
In this work, we propose a novel SGG-HT framework to generate an unbiased scene graph, as described below.

\vspace{-5pt}
\subsection{Curriculum Re-weight Mechanism}
\label{sec:CRM}

Many methods~\cite{loss:focalloss,loss:cbloss, undersamp2} are proposed to solve the imbalanced data distribution problem, but they can't achieve satisfying performance in SGG.  As mentioned in Sec.~\ref{sec:intro}, handling the imbalanced issue requires the robust features of head predicates at first. 
% With the robust features of head predicates, tail predicates can be learned quickly and achieve superior performances. 
To this end, we introduce the Curriculum Re-weight Mechanism, which adjusts the relative weights between head and tail predicates with the progress of training through a Curriculum Decay Factor \textbf{$\lambda$}, rather than assigning large weights to the tail and minor weights to the head directly. 
% Based on the $L_{CE}$, we introduce the Dynamic Decay Factor \textbf{$\lambda$},
% \gyy{comma can not connect two sentences.} 
% It adjusts the relative weight between head and tail predicates with the progress of training, achieves the learning from easy to hard\gyy{what's mean???}.\gyy{split these sentences for clear.} 
The refined loss function is as follows:
\vspace{-7pt}
\begin{equation}\label{Eq:3.2}
\vspace{-5pt}
  \mathcal{L}_{CRW}(z,y) =  - \sum_{i=1}^{R+1}(\lambda_i w_i) y_i \log{\frac{e^{z_i}}{\sum_{j=1}^{R+1} e^{z_j}}},
\end{equation}
where $w_i$ is the weight of class $i$ computed with a state-of-the-art re-weighting method~\cite{loss:cbloss}. The Curriculum Decay Factor \textbf{$\lambda_i$} is defined as:
\vspace{-5pt}
\begin{equation}
\lambda_i=
\begin{cases}
max(\varphi(l), \alpha)& \text{if $i \in H$}\\
1& \text{otherwise}
\end{cases},
\vspace{-3pt}
\end{equation}
where $H$ is the set of head predicate indexes selected by the number of predicate samples. $\varphi(l)$ is a decreasing function from 1 to 0 that represents the weights allocated to the head predicates. In order to prevent the forgetting of head predicates, a threshold hyperparameter $\alpha$ is used to avoid zero weights for them.
The $\varphi(l)$ is defined as:
\vspace{-3pt}
\begin{equation}\label{eql-4}
  \varphi(l) = 1 - \frac{l}{L},
  \vspace{-3pt}
\end{equation}
where $l$ is the current training iteration, and $L$ refers to the total training iterations. It is a linear function. 

% The class-generic feature of head feature is simple,   Through the Dynamic Transfer Factor $\lambda$, the model can gradually focus on the learning of class-specific features of the tail predicates.  

% 英文中没有总分总的结构，所以要把下面这段融入到行文中

% The samples of head predicates are sufficient, and their features are fundamental and easy to learn.
% With the robust features of head predicates, tail predicates can be learned quickly and achieve satisfying performances.
% Thus, we learn these head predicates at first and then pay more attention to the tail ones, rather than assigning a large weights to the tail and a minor value to the head at the whole training process.
% In order to prevent the forgetting of head predicates, a threshold hyperparameter $\alpha$ is used to to avoid zero weights for them.

% we gradually reduce the weight of head predicates and then pay more attention to learn the class-specific features of tail predicates
% 通过这个调节因子的控制，模型能够将注意力逐渐的集中于尾部谓词上面。 由于头部谓词的数量很多，因此其能够
% 出发点：目前的方法都没有注意到头部谓词和尾部谓词之间的特征相似性以及头部谓词的特征是简单且可以共享的.  基于此，本文提出一个XXX学习机制，其能够
% 头部谓词的类通用特征是简单的，而且其样本丰富。因此可以随着学习的进行逐渐降低其关注度，重点的去学习尾部谓词的基于类通用特征的类特定特征。为了防止对于头部谓词学习的遗忘，所以使用了一个阈值α来对保留对其一定的关注度

\begin{table*}[htp] \centering 
\caption{Compared our method SGG-HT with various state-of-the-art methods. $\diamond$ denotes the implementation in ~\cite{sgg:sggbenchmark}.}
\resizebox{\textwidth}{!}{
\begin{tabular}{cccccccccc}
\hline
\multirow{2}{*}{Models} & \multicolumn{3}{c}{PredCls} & \multicolumn{3}{c}{SGCls} & \multicolumn{3}{c}{SGDet} \\
\cline{2-10}
& mR@20 & mR@50 & mR@100 & mR@20 & mR@50 & mR@100 & mR@20 & mR@50 & mR@100 \\
\hline
Motifs$^\diamond$~\cite{sgg:motifs, sgg:sggbenchmark} & 11.67 & 14.79 & 16.08 & 6.68 & 8.28 & 8.81 & 4.98 & 6.75 & 7.90 \\
VCTree$^\diamond$~\cite{sgg:vctree, sgg:sggbenchmark} & 13.12 & 16.74 & 18.16 & 9.59 & 11.81 & 12.52 & 5.38 & 7.44 & 8.66 \\ 
Transformer$^\diamond$~\cite{sgg:sggbenchmark} & 12.77 & 16.30 & 17.63 & 8.14 & 10.09 & 10.73 & 6.01 & 8.13 & 9.56\\
PCPL~\cite{sgg:pcpl} & - & 35.2 & 37.8 & - & 18.6 & 19.6 & - & 9.5 & 11.7 \\ 
% BGNN~\cite{sgg:bgnn} & - & 30.4 & 32.9 & - & 14.3 & 16.5 & - & 10.7 & 12.6\\
Motifs (TDE)~\cite{sgg:tde} & 18.5 & 24.9 & 28.3 & 11.1 & 13.9 & 15.2 & 6.6 & 8.5 & 9.9\\ 
VCTree (TDE)~\cite{sgg:tde} & 18.4 & 25.4 & 28.7 & 8.9 & 12.2 & 14.0 & 6.9 & 9.3 & 11.1 \\ 
% VTransE (TDE)~\cite{sgg:tde} & 18.9 & 25.3 & 28.4 & 9.8 & 13.1 & 14.7 & 6.0 & 8.5 & 10.2 \\
Motifs (CogTree)~\cite{sgg:cogtree} & 20.9 & 26.4 & 29.0 & 12.1 & 14.9 & 16.1 & 7.9 & 10.4 & 11.8 \\ 
SG-Transformer (CogTree)~\cite{sgg:cogtree} & 22.9 & 28.4 & 31.0 & 13.0 & 15.7 & 16.7 & 7.9 & 11.1 & 12.7 \\ 
VCTree (CogTree)~\cite{sgg:cogtree} & 22.0 & 27.6 & 29.7 & 15.4 & 18.8 & 19.9 & 7.8 & 10.4 & 12.1 \\ 
% Motifs (EBM)~\cite{sgg:ebm} & 14.17 & 18.02 & 19.53 & 8.18 & 10.22 & 10.98 & 5.66 & 7.72 & 9.27 \\
% VCTree (EBM)~\cite{sgg:ebm} & 14.2 & 18.19 & 19.72 & 10.4 & 12.54 & 13.45 & 5.67 & 7.71 & 9.1 \\
% VCTree-TDE (EBM)~\cite{sgg:ebm} & 19.87 & 26.66 & 29.97 & 13.86 & 18.2 & 20.45 & 7.1 & 9.69 & 11.6 \\
Transformer (BA-SGG)~\cite{sgg:ba-sgg} & 26.7 & 31.9 & 34.2 & 15.7 & 18.5 & 19.4 & 11.4 & 14.8 & 17.1\\ 
Motifs (BA-SGG)~\cite{sgg:ba-sgg} & 24.8 & 29.7 & 31.7 & 14.0 & 16.5 & 17.5 & 10.7 & 13.5 & 15.6 \\ 
VCTree (BA-SGG)~\cite{sgg:ba-sgg} & 26.2 & 30.6 & 32.6 & 17.2 & 20.1 & 21.2 & 10.6 & 13.5 & 15.7\\ 
% Transformer (Re-weight*) & 30.07 & 35.52 & 37.88 & 18.23 & 21.09 & 22.13 & 12.53 & 16.32 & 19.59 \\ 
\hline \hline

\textbf{Transformer (SGG-HT)} & \textbf{34.52} & \textbf{40.28} & \textbf{42.60} & \textbf{18.86} & \textbf{22.36} & \textbf{24.71} & \textbf{13.73} & \textbf{17.68} & \textbf{20.64} \\
\textbf{Motifs (SGG-HT)} & \textbf{32.09} & \textbf{38.01} & \textbf{39.43} & \textbf{19.35} & \textbf{22.43} & \textbf{23.42} & \textbf{13.10} & \textbf{17.21} & \textbf{20.19} \\ 
\textbf{VCTree (SGG-HT)} & \textbf{32.28} & \textbf{38.32} & \textbf{40.21} & \textbf{22.36} & \textbf{25.23} & \textbf{27.11} & \textbf{12.37} & \textbf{16.05} & \textbf{18.36} \\ 
\hline
\end{tabular}
}
\vspace{-10pt}
\label{tab:mR@k} 
\end{table*}

\vspace{-2pt}
\subsection{Semantic Context Module}
\label{sec:SCM}
% 该模块功能：1）为了维持场景图的语义和真实场景图语义的一致性 2）探索关系三元组之间的相关性
% 现存的SGG方法都是使用一个分类器对关系单独的进行分类，没有探索分类后生成的场景图与真实场景图之间语义的一致性以及生成的场景图中三元组彼此之间的相关性。
% 为了缓解经验预测，我们提出了ST that使用语义信息来确保一致性

After the prediction of predicates, we get a set of predicate probabilities $\{p_1, p_2,...,p_N\}$, where $N$ is the total number of relations in an image and $p_i \in \mathbb{R}^{R+1}$.
% Next, each predicate probability is mapped to one 200-dimensional vector as predicate semantic representation with the pre-trained word embedding model. 
Next, we obtain the predicate semantic representations $\{s_1^p, s_2^p,..., s_N^p\}$ by mapping each predicate probability to a 200 dimensional vector with a pre-trained word embedding model (GloVe).
Then, the semantic representations of the subject and object are concatenated with the corresponding predicate semantic representation to get the relation triplet semantic representation, as follows:
\vspace{-1pt}
\begin{equation}
    s_i^r = Concat([s_i^s; s_i^p; s_i^o])W, W \in \mathbb{R}^{600 \times D}.
    \vspace{-1pt}
\label{equ.sc}
\end{equation}
where $W$ is a trainable linear projection that maps the concatenated semantic representation to $D$ dimensions. In addition, we add a global node $s_{global}$ as the global semantic representation of the whole graph, defined as follows:
\vspace{-3pt}
\begin{equation}
\vspace{-3pt}
    s_{global} = \frac{1}{N} \sum_{i=1}^{N}{s_i^r}.
\end{equation}
The same processing is performed on the ground truth relation triplets to get $t_i^r$ and $t_{global}$. Then, the conventional Transformer~\cite{transformer} encoder is used to construct contextual semantic representations for relation triplets. For simplicity, we denote the conventional Transformer as \textbf{$Trans(\cdot)$}, where the queries $Q$, keys $K$ and values $V$ share the same input.  The input of $Trans(\cdot)$ is  $S^r=\{s_1^r,s_2^r,...,s_N^r,s_{global}\}$ and then the contextual semantic representation, $\widetilde{S^r}$, is computed as follows:
\vspace{-2pt}
\begin{equation}
\vspace{-2pt}
    \widetilde{S^r} = Trans(S^r),
\end{equation}
where $\widetilde{S^r}=\{\widetilde{s^r_1},\widetilde{s_2^r},...,\widetilde{s_N^r},\widetilde{s}_{global}\}$. 
The ground truth contextual semantic representation $\widetilde{T^r}$ is also obtained with  $Trans(\cdot)$. Then, the $\widetilde{s}_{global}$ and $\widetilde{t}_{global}$ are used to compute the semantic gap, and a mean-squared loss is used to minimize it:
\vspace{-5pt}
\begin{equation}
    \mathcal{L}_{SC} = \frac{1}{D} {\Vert\widetilde{s}_{global} - \widetilde{t}_{global}\Vert}^2,
\label{eq.sc_loss}
\end{equation}
where $D$ is the same as in Eq.~(\ref{equ.sc}).  
Besides, $\{\widetilde{s^r_1},\widetilde{s_2^r},...,\widetilde{s_N^r}\}$ are used for predicate classification to obtain refined predicate logits $\widetilde{z}$. Afterward, the refined predicate logits combine the original predicate logits $z'$ to get the finial predicate logits, as follows: 
\vspace{-4pt}
\begin{equation}
\vspace{-3pt}
    z = z'  + \widetilde{z}.
\end{equation}
It can correct the prediction errors from the semantic perspective.
Finally, the total loss for training the scene graph generator is computed by:
\vspace{-4pt}
\begin{equation}
    \mathcal{L}_{total} = \mathcal{L}_{CRW} + \mathcal{L}_{SC}. 
\end{equation}
\vspace{-4pt}

\vspace{-17pt}
\section{EXPERIMENT}
\vspace{-2pt}
% In this section, we first introduce the experimental settings and some key implementation details of our methods. After that, we compare we proposed methods with state-of-the-art scene graph generation methods. Finally, we present an ablation study on the components we proposed.

\subsection{Experimental Settings}
\textbf{Datasets.} Our models is evaluated on the widely used benchmark, Visual Genome (VG), which is composed of 108K images with 75K object categories and 37K predicate categories. Since the majority of the annotations are noisy, we follow previous works~\cite{sgg:tde, sgg:ba-sgg, sgg:cogtree} and adopt the most popular split from ~\cite{sgg:imp}, which contains the most frequent 150 object categories and 50 predicate categories. Moreover, the VG dataset is divided into a training set with 70\% of the images and a testing set with the remaining 30\% and 5K images from the training set for validation. \\
\noindent \textbf{Evaluation.} We follow previous works~\cite{sgg:ba-sgg,sgg:cogtree,sgg:graphrcnn,sgg:tde} to evaluate our method on three subtasks: (1) Predicate Classification (\textbf{PredCls}): given the ground-truth bounding boxes and object labels in an image, predict the relationship labels; (2) Scene Graph Classification (\textbf{SGCls}): given the ground-truth bounding boxes in an image, predict the object labels and the relationship labels; (3) Scene Graph Detection (\textbf{SGDet}): given an image, predict the scene graph from scratch. Due to the extremely long-tailed data distribution in VG dataset, we use the \textbf{Mean Recall@K} (mR@K) as our main evaluation metric, and the \textbf{Recall@K} is also reported briefly.

\vspace{-5pt}
\subsection{Implementation Details}
We adopt a pre-trained Faster RCNN~\cite{faster-rcnn} with ResNeXt-101-FPN~\cite{fpn, resxnet} as the backbone object detector, and freeze the model parameters during the training. The $\alpha$ in Curriculum Re-weight Mechanism is set to 0.25 and the $D$ in Semantic Context Module is set to 512. Models are trained by SGD optimizer with 30K iterations. The batch size and learning rate are set to 12 and $12 \times 10^{-3}$. Besides, we incorporate frequency bias~\cite{sgg:motifs} into the training and inference stages. Experiments are implemented with PyTorch and trained with NVIDIA TITAN XP GPUs. 

\vspace{-4pt}
\subsection{Comparison with State-of-the-art Methods}
% We evaluate our method on three models: Transformer~\cite{sgg:sggbenchmark}, Motifs~\cite{sgg:motifs} and VCTree~\cite{sgg:vctree}.
We compare our method with state-of-the-art methods.
The comparison results are summarized in Tab.~\ref{tab:mR@k}.
The results show that our method achieves the best performance in all evaluation metrics among all the comparison methods, reaching 40.28 mR@50 for PredCls, 22.36 mR@50 for SGCls and 17.68 mR@50 for SGDet with Transformer. 
As for the Recall@K metric, we compare our method with debiasing methods CogTree, TDE and PCPL on the task of PredCls, and the results are shown in Tab.~\ref{tab:R@K}. Our method also outperforms them with this metric because they pay too much attention to the tail predicates.
Overall, these results demonstrate that our method can maintain the performance of the head predicates while significantly improving the performance of the tail predicates. 

\begin{table}[t]
    \caption{State-of-the-art comparison on R@K.}
    \centering
    \begin{tabular}{c|ccc}
    \hline
     \multirow{2}{*}{Method} & \multicolumn{3}{c}{PredCls} \\
    \cline{2-4}
    & R@20 & R@50 & R@100 \\ 
    \hline
    Motifs (CogTree)~\cite{sgg:cogtree} & 31.1 & 35.6 & 36.8 \\
    VCTree (CogTree)~\cite{sgg:cogtree} & 39.0 & 44.0 & 45.4 \\ 
    Motifs (TDE)~\cite{sgg:tde} & 33.6 & 46.2 & 51.4 \\
    VCTree (TDE)~\cite{sgg:tde} & 36.2 & 47.2 & 51.6 \\
    PCPL~\cite{sgg:pcpl} & - & 50.8 & 52.6 \\ 
    \hline \hline
    \textbf{Motifs (SGG-HT)} & \textbf{42.41} & 49.75 & 51.88 \\
    \textbf{VCTree (SGG-HT)} & \textbf{43.51} & \textbf{50.85} & \textbf{52.94} \\
    \hline
    \end{tabular}
    \label{tab:R@K}
    \vspace{-13pt}
\end{table}

\begin{table}[t]

    \caption{Ablation study for the proposed components.}
    \centering
    \setlength{\tabcolsep}{1.5mm}{
    \begin{tabular}{c|cc|ccc}
    
    \hline
    \multirow{2}{*}{Exp}& \multicolumn{2}{c|}{Method} & \multicolumn{3}{c}{PredCls} \\
    \cline{2-6}
    & CRM & SCM & mR@20 & mR@50 & mR@100 \\ 
    \hline
    1 & - & - & 30.07 & 35.52 & 37.88 \\
    2 & $\surd$ & - & 33.21$\uparrow$\textbf{3.14} & 38.97$\uparrow$\textbf{3.45} & 41.23$\uparrow$\textbf{3.35} \\ 
    3 & - & $\surd$ & 32.04$\uparrow$\textbf{1.97} & 38.05$\uparrow$\textbf{2.53} & 40.14$\uparrow$\textbf{2.26} \\
    4 & $\surd$ & $\surd$ & 34.52$\uparrow$\textbf{4.45} & 40.28$\uparrow$\textbf{4.76} & 42.60$\uparrow$\textbf{4.72} \\
    \hline
    \end{tabular}}
    \label{tab:abeach}
    \vspace{-7pt}
\end{table}

\vspace{-5pt}
\subsection{Ablation Study}
% To investigate each component's contribution and possible variants in our proposed SGG-HT framework, we construct extensive experiments.  
Extensive experiments are conducted to investigate each component’s contribution and possible variants in our proposed SGG-HT framework. 
Moreover, we use Transformer with re-weighting~\cite{loss:cbloss} as the baseline and only perform the task of PredCls for fast validation.\\
\noindent \textbf{Effectiveness of CRM and SCM}.
An ablation study is performed to validate the effectiveness of Curriculum Re-weight Mechanism (CRM) and Semantic Context Module (SCM). Results are shown in Tab.~\ref{tab:abeach}. There are the results of four experiments in the table. Exp (1): Baseline, Transformer trained with re-weighting~\cite{loss:cbloss}. Exp (2): CRM is added to the baseline to learn from head to tail. Exp (3): SCM is used to relieve semantic deviation. Exp (4): CRM and SCM are combined as mentioned in Sec.~\ref{sec.method}. 
Compared with baseline, CRM and SCM increase the mR@50 metric by 3.45 and 2.53,  respectively.
Moreover, combining CRM and SCM, our method further improves the performance, demonstrating the effectiveness of our SGG-HT framework.

% For Exp 2-3, a consistent performance improvement when equipping with our components can be seen, demonstrating the effectiveness of individual components in our method.
% \gyy{ADVICE give Give specific score when adding CRS and ST, respectively and explain the results}
% CRS提升了3.45个点，说明了我们CRS

\begin{figure}[t]
  \center
  \includegraphics[width=1\linewidth]{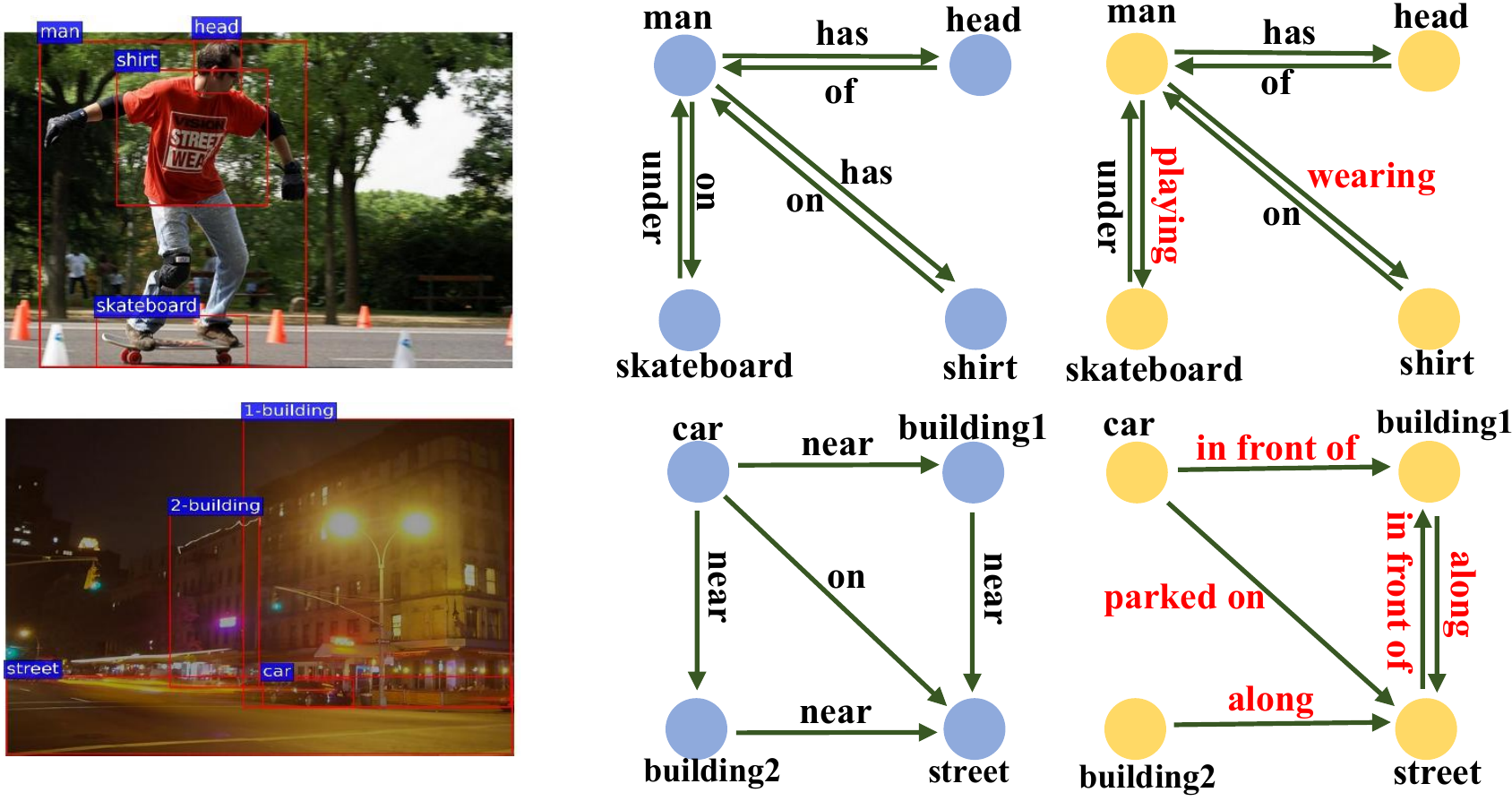}
  \vspace{-22pt}
\caption{Qualitative Results. Visualization results of VCTree in blue and VCTree (SGG-HT) in yellow on the PredCls task. The red words represent the fine-grained predicates.  
}
\vspace{-13pt}
\label{fig:fig3}
\end{figure}

\begin{table}[!t]
    \caption{Ablation study for CRM.}
    \centering
    \begin{tabular}{c|c|ccc}
    \hline
    \multirow{2}{*}{Exp}&  \multirow{2}{*}{Func} & \multicolumn{3}{c}{PredCls} \\
    \cline{3-5}
    & & mR@20 & mR@50 & mR@100 \\ 
    \hline
    1 & Exp & 33.81 & 39.70 & 41.88 \\
    2 & Cos & \textbf{34.55} & 40.13 & 42.45 \\
    3 & Linear & 34.52 & \textbf{40.28} & \textbf{42.60} \\
    \hline
    \end{tabular}
    \label{tab:ab1}
    \vspace{-13pt}
\end{table}

\begin{table}[!t]
    \caption{Ablation study for SCM.}
    \centering
    \begin{tabular}{c|c|ccc}
    \hline
    \multirow{2}{*}{Exp}&  \multirow{2}{*}{Method} & \multicolumn{3}{c}{PredCls} \\
    \cline{3-5}
    & & mR@20 & mR@50 & mR@100 \\ 
    \hline
    1 & Mean & 33.66 & 39.97 & 42.54 \\
    2 & Global & \textbf{34.52} & \textbf{40.28} & \textbf{42.60} \\
    \hline
    \end{tabular}
    \vspace{-7pt}
    \label{tab:ab2}
\end{table}

\noindent \textbf{Variants to Curriculum Re-weight Mechanism}. For Curriculum Re-weight Mechanism, we investigate the variants of $\varphi(l)$ mentioned in Eq.~(\ref{eql-4}) and compare the linear function with the other two decreasing functions as follows:
\begin{enumerate}
\vspace{-3pt}
    \item[(1)] Exponential function, indicates the speed of transfer from fast to slow, defined as:\vspace{-12pt}
    
    \begin{equation}
    \vspace{-5pt}
        \varphi(l) = \nu^{\frac{l}{L}}, \text{($0 < \nu < 1$)}.
    \end{equation}
    \item[(2)] Cosine function, indicates the speed of transfer from slow to fast, defined as:
    \vspace{-10pt}
    \begin{equation}
    \vspace{-10pt}
        \varphi(l) = \cos(\frac{\pi}{2} \times \frac{l}{L}).
    \end{equation}
\end{enumerate}
The experimental results are shown in Tab.~\ref{tab:ab1}. These results show that the liner function outperforms the cosine function in mR@50 and mR@100 and achieves better performance than the exponential function in all metrics. Therefore, in the SGG-HT framework, the Linear function is utilized to decay the weights on the head predicates.

\noindent \textbf{Variants to Semantic Context Module}. In this module, a global node is utilized as $s_{global}$ (Global) mentioned in Sec.~\ref{sec:SCM} for the semantic representation of the whole graph. In order to demonstrate the effectiveness of $s_{global}$, we first remove $s_{global}$ in $S^r$, and then take the average of $\{\widetilde{s_1^r},\widetilde{s_2^r},...,\widetilde{s_N^r}\}$ (Mean) as $\widetilde{s}_{global}$ in Eq.~(\ref{eq.sc_loss}). 
Shown in Tab.~\ref{tab:ab2}, Global exceeds Mean in all metrics, which shows the superiority of the global node $s_{global}$. 
It indicates that the Global fully considers the semantic context correlations among all relation triplets, which better represents the semantics of the whole scene graph, compared with the Mean.   

% \gyy{Give the reason???why global is better than mean}

% global可以从全局的角度来综合性的表述场景图的语义，而mean只能局部的平均， 而且从图中可以看出，

\subsection{Qualitative Results}
Some scene graph generation visualization results of VCTree and VCTree (SGG-HT) are in Fig.~\ref{fig:fig3}. It is obvious that VCTree (SGG-HT) generates more fine-grained relationships than plain VCTree, such as ``man wearing shirt", ``man playing skateboard", ``car parked on street" and ``street in front of building1". These results demonstrate the effectiveness of our method for the balanced scene graph generation.

\section{CONCLUSION}

% we first reveal previous debiasing methods ignore the importance of the head predicates. To address this problem, 
In this work, for learning robustness scene graphs, we propose a novel framework SGG-HT, which contains a Curriculum Re-weight Mechanism that regulates the model to learn to generate scene graphs from head to tail, and a Semantic Context Module that alleviates the semantic deviation by ensuring the semantic consistency between the generated scene graph and the ground truth. Extensive experiments show that our SGG-HT framework significantly improves the performance of scene graph generation and achieves the new state-of-the-art performance.

% \section{Citations and References}

% List and number all bibliographical references at the end of the paper. The references can be numbered in alphabetic order or in order of appearance in the document. When referring to them in the text, type the corresponding reference number in square brackets as shown at the end of this sentence~\cite{Morgan2005}. All citations must be adhered to IEEE format and style. Examples such as~\cite{Morgan2005},~\cite{cooley65} and~\cite{haykin02} are given in Section 12.


\bibliographystyle{IEEEbib}
\bibliography{icme2022template}

\end{document}